# Modifying Bayesian Networks by Probability Constraints

**Yun Peng**
Department of Computer Science and
Electrical Engineering
University of Maryland Baltimore County
Baltimore, MD 21250
ypeng@csee.umbc.edu

**Zhongli Ding**
Department of Computer Science and
Electrical Engineering
University of Maryland Baltimore County
Baltimore, MD 21250
zding1@csee.umbc.edu

## Abstract

This paper deals with the following problem: modify a Bayesian network to satisfy a given set of probability constraints by only change its conditional probability tables, and the probability distribution of the resulting network should be as close as possible to that of the original network. We propose to solve this problem by extending IPFP (iterative proportional fitting procedure) to probability distributions represented by Bayesian networks. The resulting algorithm E-IPFP is further developed to D-IPFP, which reduces the computational cost by decomposing a global E-IPFP into a set of smaller local E-IPFP problems. Limited analysis is provided, including the convergence proofs of the two algorithms. Computer experiments were conducted to validate the algorithms. The results are consistent with the theoretical analysis.

## 1 INTRODUCTION

Consider a Bayesian network (BN) $N$ on a set of variable $X$ that models a particular domain. $N$ defines a distribution $P(x)$. Suppose you are given a probability distribution $R(y)$ on a subset of the variables $Y \subseteq X$ and $R$ does not agree with $P$ (i.e., $P(y) \neq R(y)$). Can you change $N$ to $N'$ so that its distribution $P'(x)$ satisfies $R$? Can you do so with more than one such probability constraints $R_1(y^1), R_2(y^2), \cdots R_m(y^m)$? Moreover, can you do so without changing the structure of $N'$ (i.e., only modifying CPTs, the conditional probability tables of $N$)?

Problems of this kind can be found in designing new BNs, merging small BNs into a large one, or refining an existing one with new or more reliable probability information. For example, when designing a BN for heart disease diagnosis, it is relatively easy to obtain a consensus among domain experts on what factors affect heart diseases and how they are causally related to one another. This knowledge of qualitative associations can then be used to define the networks structure, i.e., the directed acyclic graph (DAG) of the BN.

However, it is not that easy to obtain the conditional probability tables for each of the variables. Experts' opinions are often coarse (the likelihood of $a$ causes $b$ is "high" but that of $c$ cause $b$ is "low"), not in a uniform scale ("high" given for one association may not mean exactly the same for another association), not in the form of CPT (not the likelihood of causing $b$ by the combination of all possible states of $a$ and $c$). Learning CPT from statistical data is also problematic. Most learning methods require samples of complete instantiations of all variables, but in many real word applications, especially those involving a large number of variables, statistical data are fragmented, represented by, say, a number of low-dimensional distributions over subsets of the variables. In the heart disease example, one may obtain a distribution of *drinking* and *heart diseases* from a survey concerning effects of drinking on people's health, and a distribution of *smoking* and *heart diseases* from a survey concerning effects of smoking on people's health. But none of them include both *drinking* and *smoking*, two of the important causal factors to *heart diseases*. Moreover, a new survey on drinking with larger samples and improved survey methods may give a more accurate distribution of *drinking* and *heart diseases*, the BN needs to adapt itself to the new data.

Iterative proportional fitting procedure (IPFP) is a mathematical procedure that iteratively modifies a probability distribution to satisfy a set of probability constraints while maintaining minimum Kulback-Leibler distance to the original distribution. One would think this kind of BN modification tasks can be accomplished by applying IPFP on the distribution of the given BN. This approach will not work well for at least two reasons. First, theoretically the distribution resulted from the IPFP process, although satisfying all the constraints, may not always be consistent with the interdependencies imposed by the network structure, and thus cannot be used to generate new CPTs properly. Secondly, since IPFP works on the joint distribution of all variables of the BN, it becomes computational intractable with large BNs.

In this paper, we describe our approach to address both of these problems. The first problem is resolved by algorithm E-IPFP, which extends IPFP by converting the structural invariance to a new probability constraint. The second problem is resolved by algorithm D-IPFP. This algorithm decomposes a global E-IPFP into a set of smaller, local E-IPFP problems, each of which corresponds to one constraint and only involves variables that are relevant to those in that constraint.

The rest of this paper is organized as follows. Section 2 states precisely the BN modification problems we intend to solve. Section 3 gives a brief introduction to IPFP. E-IPFP and its convergence proof are given in Section 4. Section 5 describes D-IPFP and shows that a significant saving is achieved with reasonable relaxation of the minimum distance requirement. Convergence proof of the algorithm is also given. Computer experiments of limited scope were conducted to validate the algorithms and to give us a sense of how expensive this approach may be. Experiment results are given in Section 6. Section 7 concludes this paper with comments on related works and suggestions for future research.

## 2   THE PROBLEM

We adopt the following notation in this paper. A BN is denoted as $N$, different BN's are differentiated by subscripts. $G$ denoted the structure (i.e., the DAG) of BN, and $C$ for the set of conditional probability tables (CPTs) of $N$. $X$, $Y$, $Z$, ..., are for a sets of variables, and $x$ an instantiation of $X$. Individual variables are indicated by subscripts, for example, $X_i$ is a variable in $X$ and $x_i$ its instantiation. Capital letters $P$, $Q$, $R$, are for probability distributions. A *probability constraint* $R_i(y)$ to distribution $P(x)$ is a distribution on $Y \subseteq X$. $P(x)$ is said to satisfy $R_i(y)$ if $P(y) = R_i(y)$. $R$ denotes a set of constraints $R_i(y)$.

We say a network $N_0$ on $X = \{X_1, X_2, \cdots, X_n\}$ has DAG $G_0$ and CPT set $C_0$ where each CPT in $C_0$ is in the form of $P(x_i | \pi_i)$ where $\pi_i$ is the set of parents of $X_i$ as specified in $G_0$. Also, we call $P_0(x) = \Pi_{i=1}^{n} P_0(x_i | \pi_i)$ the (probability) distribution of $N_0$.

We call $P(x_i | \pi_i)$ a CPT *extracted* from $P(x)$ according to $G$ if $\pi_i$ is determined by $G$. Extraction of $P(x_i | \pi_i)$ can be done by computing $P(\pi_i)$ and $P(x_i, \pi_i)$ from $P(x)$ through marginalization or any other methods. When $P(x)$ and $G$ are given, CPT extraction is unique. When all CPTs of $N$ are replaced by those extracted from an arbitrary $P(x)$ according to $G$, its distribution $P'(x) = \Pi_{i=1}^{n} P(x_i | \pi_i)$ may not equal to $P(x)$ even though the conditional distribution of $x_i$, given $\pi_i$ are the same in both $P$ and $P'$. This is because certain conditional independences of $P'$, dictated by the network structure, does not hold for $P$.

A distribution $P(x)$ is said to be (structurally) *consistent* with $G$ of $N$ if there exists a set of CPTs $C = \{Q(x_i | \pi_i)\}$ such that $P(x) = \Pi_{i=1}^{n} Q(x_i | \pi_i)$. Since when $G$ is given, $P(x)$ uniquely determines $P(x_i | \pi_i) \forall i$, so if $P(x)$ is consistent with $G$ then $P(x) = \Pi_{i=1}^{n} P(x_i | \pi_i)$. Consequently, if both $P$ and $P'$ are consistent with $G$, then $P = P'$ if and only if $C = C'$.

We use *I-divergence* (also known as *Kullback-Leibler distance* or *cross-entropy*) to measure the distance between two distributions $P$ and $Q$ over $X$:

$$I(P \| Q) = \begin{cases} \sum_{P(x)>0} P(x) \log \frac{P(x)}{Q(x)} & \text{if } P \ll Q \\ +\infty & \text{otherwise} \end{cases} \quad (1)$$

where $P \ll Q$ means $Q$ dominates $P$ (i.e., $\{x | P(x) > 0\} \subseteq \{x' | Q(x') > 0\}$). $I(P \| Q) \geq 0$ for all $P$ and $Q$, the equality holds only if $P = Q$.

We say $P(x)$ is an *I-projection* of $Q(x)$ on a set of constraints $R$ if $I(P \| Q)$ is smallest among all distributions that satisfy $R$.

With the above notation, we can state precisely the problem we are trying to solve: for a given $N$ over variables $X$ with distribution $Q$ and a set of consistent constraints $R$, find $N^*$ that meets the following three requirements:

(a) $G = G^*$ (both networks have the same structure);
(b) $Q^*$, the distribution of $N^*$, satisfies all constraints in $R$; and
(c) $I(Q^* \| Q)$ is minimum among all distributions that meet requirements (a) and (b).

## 3   BRIEF INTRODUCTION TO IPFP

Iterative proportional fitting procedure (IPFP) was first published in (Kruithof 1937). Shortly after, it was proposed as a procedure to estimate cell frequencies in contingency tables under some marginal constraints (Deming and Stephan 1940). Csiszar (1975) provided an IPFP convergence proof based on *I-divergence* geometry. Vomel rewrote a discrete version of this proof in his PhD thesis (Vomlel 1999). IPFP was extended in (Bock 1989, Cramer 2000) as conditional iterative proportional fitting procedure (CIPFP) to also take conditional distributions as constraints, and the convergence was established for the finite discrete case.

For a given distribution $Q_0(x)$ and a set of consistent constraints $R$, IPFP converges to $Q^*(x)$ which is an I-projection of $Q_0$ on $R$ (assuming there exists at least one distribution that satisfies $R$). This is done by iteratively modifying the distributions according to the following formula, each time using one constraint in $R$:

$$Q_k(x) = \begin{cases} 0 & \text{if } Q_{k-1}(y) = 0 \\ Q_{k-1}(x) \cdot \frac{R_i(y)}{Q_{k-1}(y)} & \text{otherwise} \end{cases}$$

where $m$ is the number of constraints in $R$, and $i = ((k-1) \mod m) + 1$ determines the constraint used at step $k$. For clarity, in the rest of this paper, we write the above formula as

$$Q_k(x) = Q_{k-1}(x) \cdot \frac{R_i(y)}{Q_{k-1}(y)} \quad (2)$$

with the understanding that $Q_k(x) = 0$ if $Q_{k-1}(y) = 0$. What (2) does at step k is to change $Q_{k-1}(x)$ to $Q_k(x)$ so that $Q_k(y) = R_i(y)$.

For a given $N_0$ and its distribution $Q_0(x) = \Pi_{i=1}^{n} Q_0(x_i | \pi_i)$ and constraint set $R$, one can always obtain an I-projection $Q^*(x)$ of $Q_0(x)$ on $R$ by IPFP. However, $Q^*(x)$ is not guaranteed to be consistent with $G_0$. This is especially true if some constraints involve more than one variables and they span over more than one CPT. This problem is illustrated in the following examples with a small network $N$ of variables $\{A, B, C, D\}$ depicted in Figure 1 below. Figures 2 and 3 give the results of applying IPFP on $N$ with constraint sets $\{R_1(B), R_2(C)\}$ and $\{R_3(A, D)\}$, respectively.

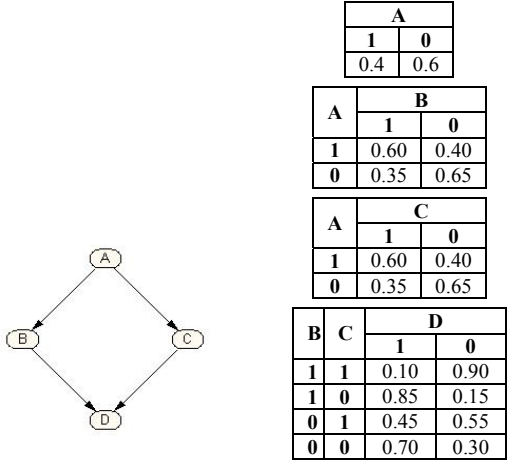

Figure 1: Network $N$ of $X = \{A, B, C, D\}$ and its CPTs

Figure 2 below gives $Q^*(a, b, c, d)$, the distribution resulted from IPFP with two constraints $R_1(b) = (0.61, 0.39)$ and $R_2(c) = (0.83, 0.17)$. I-divergence of $Q^*$ to the original distribution is **0.5708**. Also given are the CPTs of the four variables extracted from $Q^*$ according to the network structure. We can verify that 1) $Q^*(b) = (R_1(b)$ and $Q^*(c) = R_2(c)$ (i.e., $Q^*$ satisfies both constraints), and 2) $Q^*(a, b, c, d) = Q^*(a) \cdot Q^*(b|a) \cdot Q^*(c|a) \cdot Q^*(d|b,c)$ (i.e., $Q^*$ is consistent with the network structure). Note here that CPTs of three nodes ($A, B, C$) have been changed.

However, it is different when a single constraint $R_3(a, d) = (0.1868, 0.2132, 0.1314, 0.4686)$ is used. As can be seen in Figure 3, the resulting distribution, although satisfying $R_3$, is *not* consistent with the structure of $N$. This is because $A$ and $D$ are not within a single CPT. Its I-divergence to the original distribution is **0.2611**.

| A | |
|---|---|
| 1 | 0 |
| 0.4083 | 0.5916 |

| A | B | |
|---|---|---|
| | 1 | 0 |
| 1 | 0.3677 | 0.6323 |
| 0 | 0.7772 | 0.2227 |

| A | C | |
|---|---|---|
| | 1 | 0 |
| 1 | 0.9066 | 0.0933 |
| 0 | 0.7771 | 0.2220 |

| B | C | D | |
|---|---|---|---|
| | | 1 | 0 |
| 1 | 1 | 0.10 | 0.90 |
| 1 | 0 | 0.85 | 0.15 |
| 0 | 1 | 0.45 | 0.55 |
| 0 | 0 | 0.70 | 0.30 |

| Variables | | | | Prob. |
|---|---|---|---|---|
| A | B | C | D | |
| 1 | 1 | 1 | 1 | 0.0136 |
| 1 | 1 | 1 | 0 | 0.1225 |
| 1 | 1 | 0 | 1 | 0.0119 |
| 1 | 1 | 0 | 0 | 0.0021 |
| 1 | 0 | 1 | 1 | 0.1054 |
| 1 | 0 | 1 | 0 | 0.1288 |
| 1 | 0 | 0 | 1 | 0.0169 |
| 1 | 0 | 0 | 0 | 0.0072 |
| 0 | 1 | 1 | 1 | 0.0357 |
| 0 | 1 | 1 | 0 | 0.3216 |
| 0 | 1 | 0 | 1 | 0.0871 |
| 0 | 1 | 0 | 0 | 0.0154 |
| 0 | 0 | 1 | 1 | 0.0461 |
| 0 | 0 | 1 | 0 | 0.0563 |
| 0 | 0 | 0 | 1 | 0.0206 |
| 0 | 0 | 0 | 0 | 0.0088 |

Figure 2: Distribution Modified with $R_1(B)$ and $R_2(C)$

| A | |
|---|---|
| 1 | 0 |
| 0.4 | 0.6 |

| A | B | |
|---|---|---|
| | 1 | 0 |
| 1 | 0.2051 | 0.7949 |
| 0 | 0.6094 | 0.3905 |

| A | C | |
|---|---|---|
| | 1 | 0 |
| 1 | 0.6178 | 0.3821 |
| 0 | 0.5469 | 0.4530 |

| B | C | D | |
|---|---|---|---|
| | | 1 | 0 |
| 1 | 1 | 0.0322 | 0.9677 |
| 1 | 0 | 0.5699 | 0.4300 |
| 0 | 1 | 0.3065 | 0.6934 |
| 0 | 0 | 0.4744 | 0.5255 |

| Variables | | | | Prob. |
|---|---|---|---|---|
| A | B | C | D | |
| 1 | 1 | 1 | 1 | 0.0043 |
| 1 | 1 | 1 | 0 | 0.0480 |
| 1 | 1 | 0 | 1 | 0.0244 |
| 1 | 1 | 0 | 0 | 0.0053 |
| 1 | 0 | 1 | 1 | 0.0776 |
| 1 | 0 | 1 | 0 | 0.1173 |
| 1 | 0 | 0 | 1 | 0.0805 |
| 1 | 0 | 0 | 0 | 0.0426 |
| 0 | 1 | 1 | 1 | 0.0046 |
| 0 | 1 | 1 | 0 | 0.2200 |
| 0 | 1 | 0 | 1 | 0.0729 |
| 0 | 1 | 0 | 0 | 0.0681 |
| 0 | 0 | 1 | 1 | 0.0139 |
| 0 | 0 | 1 | 0 | 0.0897 |
| 0 | 0 | 0 | 1 | 0.0400 |
| 0 | 0 | 0 | 0 | 0.0908 |

Figure 3: Distribution Modified with $R_3(A,D)$

## 4 E-IPFP

To solve the BN modification problem defined in Section 2, we first extend IPFP to handle the requirement that the solution distribution should be consistent with $G_0$, the structure of the given BN. Recall that whether a distribution $Q(x)$ is consistent with $G_0$ can be determined by if $Q(x) = \Pi_{i=1}^{n} Q(x_i | \pi_i)$, where the parent-child relation in the right hand of the equation is determined by $G_0$. We can thus treat this requirement as a probability constraint $R_r(x) = \Pi_{i=1}^{n} Q_{k-1}(x_i | \pi_i)$ in IPFP. Here $Q_{k-1}(x_i | \pi_i)$ are extracted from $Q_{k-1}(x)$ according to $G_0$. We call $R_r$ a structural constraint.

Like any other constraint $R_i$, this constraint, when applied at step k, changes $Q_{k-1}(x)$ to $Q_k(x)$. By (2) $Q_k(x) = R_r(x) = \Pi_{i=1}^{n} Q_{k-1}(x_i | \pi_i)$, thus meeting the structural consistency requirement.

## 4.1 E-IPFP ALGORITHM

Let $Q_0(x) = \Pi_{i=1}^n Q_0(x_i | \pi_i)$ be the distribution of a given network $N_0$, and $R$ be the set of $m$ given constraints. E-IPFP is a simple extension of the standard IPFP by including the structural constraint as the $(m+1)^{th}$ constraint $R_{m+1}$. The algorithm E-IPFP is stated as follows:

**E-IPFP**( $N_0(X)$, $R = \{R_1, R_2, \cdots R_m\}$ ) {
1. $Q_0(x) = \Pi_{i=1}^n Q_0(x_i | \pi_i)$ where $Q_0(x_i | \pi_i) \in C_0$;
2. Starting with k = 1, repeat the following procedure until convergence {
   2.1. i = ((k-1) mod (m+1)) + 1;
   2.2. if i < m+1
   $$Q_k(x) = Q_{k-1}(x) \cdot \frac{R_i(y)}{Q_{k-1}(y)}$$
   2.3. else
   {
   extract $Q_{k-1}(x_i | \pi_i)$ from $Q_{k-1}(x)$ according to $G_0$;
   $Q_k(x) = \Pi_{i=1}^n Q_{k-1}(x_i | \pi_i)$;
   }
   2.4. k = k+1;
   }
3. return $N^*(X)$ with $G^* = G_0$ and $C^* = \{Q_k(x_i | \pi_i)\}$;
}

As a practical matter, convergence of E-IPFP can be determined by testing if the difference between $Q_k(x)$ and $Q_{k-1}(x)$ (by any of a number of metrics) is below some given threshold.

All constraints remain constant during the iteration process except $R_{m+1}$, which changes its value every time it is applied. Nonetheless, as a distribution, when $R_{m+1}$ is applied to $Q_{k-1}(x)$, the resulting $Q_k(x) = R_{m+1}$ is an I-projection of $Q_{k-1}(x)$ on $R_{m+1}$. This is because $Q_k(x)$ is the only distribution that satisfied $R_{m+1}$ since $R_{m+1}$ is a distribution of $x$, not of a subset of $x$. As can be seen in (Csiszar 1975, Vomlel 1999), convergence of the original IPFP is a consequence of the property that each iteration of IPFP produces an I-projection of the previous distribution on a constraint. Since this condition holds for our E-IPFP, the process converges to a distribution $Q^*(x)$, and $Q^*(x)$ is an I-projection of $Q_0(x)$ on $\{R_1, R_2, \cdots R_m, R_{m+1}\}$. Since $Q^*(x)$ satisfies $R_{m+1}$, we have $Q^*(x) = \Pi_{i=1}^n Q^*(x_i | \pi_i)$, so it also satisfies the structural consistency requirement. Therefore, among those distributions that satisfy given constraints in $R$ and are consistent with $G_0$, $Q^*(x)$ has the minimum I-divergence to $Q_0(x)$.

Application of E-IPFP to the network $N$ of Figure 1 with a single constraint $R_3(A,D)$ = (0.1868, 0.2132, 0.1314, 0.4686) converges to a distribution. Comparing with the result in Figure 3 (using standard IPFP), this distribution not only satisfies $R_3$, it is also structurally consistent with $N$. However, its I-divergence to the original distribution increases slightly in absolute value (from 0.2611 in Figure 3 to **0.4419**).

## 5 D-IPFP

As can be seen in (2), the computation of both IPFP and E-IPFP is on the entire joint distribution of $X$ at every iteration. This distribution becomes prohibitively large with large $n$, making the process computationally intractable for BN of large size. Roughly speaking, when $Q_{k-1}(x)$ is modified by constraint $R_i(y)$, (2) requires to check each entry in $Q_{k-1}(x)$ against every entry of $R_i(y)$ and make the update if $x$ is consistent with $y$. The cost can be roughly estimated as $O(2^n \cdot 2^{|y|})$, which is huge when $n$ is large.

Since the joint distribution of a BN is a product of distributions of much smaller size (i.e., its CPTs), the computational cost of E-IPFP could be reduced if we can utilize the interdependencies imposed on the distribution by the network structure and only update some selected CPTs. This motivates the development of algorithm D-IPFP. D-IPFP decomposes the *global* E-IPFP (the one involving all $n$ variables) into a set of *local* E-IPFP, each for one constraints $R_i(y)$, on a small subnet of $N_0$ that contains $Y$.

First we divide constraints into two types. $R_i(y)$ is said to be *local* if $Y$ contains nothing else except one variable $X_j$ and zero or more of its parents. In other words, $y = (x_j, z^j)$ where $z^j \subseteq \pi_j$. Otherwise, $R_i(y)$ is said to be *non-local*. How to deal with local and non-local constraints in D-IPFP is given in the next two subsections.

### 5.1 LOCAL CONSTRAINTS

We have proposed previously a method to reduce the IPFP computing cost for local constraints (Ding *et al* 2004). Suppose $Q_{k-1}(x) = \Pi_{i=1}^n Q_{k-1}(x_i | \pi_i)$, i.e., $Q_{k-1}(x)$ is consistent with $G_0$. Consider a local constraint $R_i(y) = R_i(x_j, z^j \subseteq \pi_j)$. Since it is a constraint only on $x_j$ and some of its parents, updating $Q_{k-1}(x)$ by $R_i(y)$ can be done by only updating $Q_{k-1}(x_j | \pi_j)$, the CPT for $x_j$, while leaving all other CPTs intact. One problem arises: since $Q_{k-1}(x_j | \pi_j)$ is an conditional distribution on $x_j$, $Q_{k-1}(x_j | \pi_j) R_i(y) / Q_{k-1}(y)$ is in general not a probability distribution, and thus cannot be used as the CPT for $x_j$ in $Q_k(x)$. This problem can be resolved by normalization. The update rule for local constraint is

$$\begin{cases} Q_k(x_j | \pi_j) = Q_{k-1}(x_j | \pi_j) \cdot \dfrac{R_i(y)}{Q_{k-1}(y)} \alpha_k \\ Q_k(x_l | \pi_l) = Q_{k-1}(x_l | \pi_l) \qquad \forall l \neq j \end{cases}. \quad (3)$$

where

$$\alpha_k = \Sigma_{x_j} Q_{k-1}(x_j | \pi_j) \frac{R_i(y)}{Q_{k-1}(y)}, \quad (4)$$

is the normalization factor. This rule leads to

$$Q_k(x) = Q_k(x_j | \pi_j) \cdot \prod_{l \neq j} Q_{k-1}(x_l | \pi_l). \quad (5)$$

Therefore $Q_k(x)$ is consistent with $G_0$, i.e., it satisfies the structural constraint. (5) can also be written as

$$Q_k(x) = Q_{k-1}(x_j | \pi_j) \cdot \frac{R_i(y)}{Q_{k-1}(y)} \cdot \alpha_k \cdot \prod_{l \neq j} Q_{k-1}(x_l | \pi_l)$$
$$= Q_{k-1}(x) \cdot \frac{R_i(y)}{Q_{k-1}(y)} \cdot \alpha_k$$

Therefore $Q_k(x)$ is not an I-project of $Q_{k-1}(x)$ on $R_i(y)$ unless $\alpha_k = 1$. This makes the convergence with rule (3) not obvious. Instead of proving the convergence here, we delay the analysis until we deal with non-local constraints in the next subsection. Analysis results are applicable to local constraints because they are special cases of non-local constraints.

Recall the example in Figure 2 where IPFP is used with two local constraints $R_1(B)$ and $R_2(C)$, three variables (B, C, and A) have their CPTs changed in the final BN. When rule (3) is applied, only tables for B and C have been changed in the final BN. Its I-divergence to the original distribution is slightly largerr than the one obtained by IPFP of (2) (increased to **0.5711** from 0.5708).

### 5.2 NON-LOCAL CONSTRAINTS

Now we generalize the idea of (3) to non-local constraints. Without loss of generality, consider one such constraint $R_i(y)$ where $y$ spans more than one CPT. Let $S = \bigcup_{x_j \in y} \pi_j \setminus y$, i.e., $S$ contains of all parent nodes in all of variables in $Y$ except those that are also in $Y$. Multiplying all CPTs for variables in $Y$, one can construct a conditional distribution

$$Q'_{k-1}(y | s) = \prod_{X_j \in Y} Q_{k-1}(x_j | \pi_j). \quad (6)$$

With (6), we defined $Q'_{k-1}(x)$ as follows,
$$Q'_{k-1}(x) = Q_{k-1}(x)$$
$$= Q'_{k-1}(y | s) \prod_{X_l \notin Y} Q_{k-1}(x_l | \pi_l). \quad (7)$$

Now $R_i(y)$ becomes local to the table $Q'_{k-1}(y | s)$, and the rule for local constraint (3) can be applied to obtain $Q'_k(y | s)$:

$$\begin{cases} Q'_k(y | s) = Q'_{k-1}(y | s) \cdot \frac{R_i(y)}{Q'_{k-1}(y)} \alpha_k \\ Q'_k(x_l | \pi_l) = Q'_{k-1}(x_l | \pi_l) \ \forall x_l \notin y. \end{cases} \quad (8)$$

CPTs for variables outside $Y$ will remain unchanged): Next, we extract $Q_k(x_j | \pi_j)$ for all $X_j \in Y$ from $Q'_k(y | s)$ by

$$Q_k(x_j | \pi_j) = Q'_k(x_j | \pi_j)$$

The process ends with

$$Q_k(x) = \prod_{X_j \in Y} Q'_k(x_j | \pi_j) \prod_{X_l \notin Y} Q_{k-1}(x_l | \pi_l) \quad (9)$$

Update of $Q_{k-1}(x)$ to $Q_k(x)$ by $R_i(y)$ can be seen to consist of three steps. 1) form $Q'_{k-1}(y | s)$ from CPTs for $X_j \in Y$ by (6); 2) update $Q'_{k-1}(y | s)$ to $Q'_k(y | s)$ by $R_i(y)$ using (8); and 3) extract $Q'_k(x_j | \pi_j)$ from $Q'_k(y | s)$ by (9). Comparing (6), (8) and (9) with Step 1, Step 2.2 and Step 2.3 in algorithm E-IPFP, this procedure of D-IPFP amounts to one iteration of a local E-IPFP on distribution $Q'_{k-1}(y | s)$.

Consider again the network $N$ of Figure 1 with a single non-local constraint $R_3(a, d)$, we have $Y = \{A, D\}$ and $S = \{B, C\}$. The new table $\hat{Q}(a, d | b, c)$ can be computed from the product of $Q(a)$ and $Q(d | b, c)$ of the original BN. For example, one entry of this table $Q(A = 1, D = 0 | B = 1, C = 1)$ to be $0.4 \cdot 0.9 = 0.36$.

Again we applied D-IPFP to the BN in Figure 1 with the non-local constraint $R_3(A, D)$. The process converged. The resulting BN satisfies the constraint $R_3$, and only CPTs for A and B have been changed. As expected, as for the I-divergence, D-IPFP would be worse that E-IPFP (it increased from 0.4419 to 0.**7827**).

The moderate sacrifice in I-divergence with D-IPFP is rewarded by a significant saving in computation. Since $R_i(y)$ is now used to modify $Q'_{k-1}(y | s)$, not $Q_{k-1}(x)$, the cost for each step is reduced from $O(2^n \cdot 2^{|y|})$ to $O(2^{|s|+|y|} \cdot 2^{|y|})$ where $O(2^{|s|+|y|})$ is the size of CPT $Q'_{k-1}(y | s)$. The saving is thus in the order of $O(2^{n-|s|+|y|})$.

Equation (3) for local constraints can easily be seen as a special case of the D-IPFP procedure described here, with $y = x_j$ and $s = \pi_j$.

Next we analyze the convergence of D-IPFP with rules (6) – (9) in the following theorem.

**Theorem**. Let $Q_{k-1}(x) = \prod_{i=1}^n Q_{k-1}(x_i | \pi_i)$ be a probability distribution over variables $X$ where each $Q_{k-1}(x_i | \pi_i)$ is the CPT for variable $X_i$ in a Bayesian network $N$ of n variables. Let $R_i(y)$ be a probability distribution over variables $Y \subseteq X$ that is consistent with the structure of $N$, i.e., there exists a distribution over $X$ that satisfies $R_i(y)$ and its DAG is the same as that of $N$'s. Then

1. The iterations of rules (6) – (9), starting with $Q_{k-1}(x)$, will converge to a distribution $Q^*(x)$;
2. $Q^*(x)$ satisfies $R_i(y)$, i.e., $Q^*(y) = R_i(y)$;
3. $Q^*(x)$ is consistent with the structure of $N$;
4. $Q^*(x)$ is not always the I-projection of $Q_{k-1}(x)$ on $R_i(y)$.

**Proof**.

*Part 1*. Note that (6) and (7) does not change the distribution, only change its representation. (9) imposes structural constraint on $Q'_k(y | s)$, as argued in Section 4 for E-IPFP, $Q_k(x)$ is thus an I-projection of $Q'_k(x)$. Now we show that with (8), $Q'_k(x)$ is an I-projection of $Q'_{k-1}(x)$. Combining (6), (7), and (8), we have

$$Q'_k(x) = Q'_k(y \mid s) \prod_{x_l \notin y} Q'_{k-1}(x_l \mid \pi_l)$$
$$= [Q'_{k-1}(y \mid s) \cdot \prod_{x_l \notin y} Q'_{k-1}(x_l \mid \pi_l)] \cdot \frac{Q'_k(y \mid s)}{Q'_{k-1}(y \mid s)},$$
$$= Q'_{k-1}(x) \frac{Q'_k(y \mid s)}{Q'_{k-1}(y \mid s)}.$$

and viewing $Q'_k(y \mid s)$ a constraint. Since each update generates a distribution that is an I-projection of the previous distribution, again, according to (Csiszar 1975, Vomlel 1999), the process converges with
$$\lim_{k \to \infty} Q_k(x) = Q^*(x).$$

*Part 2.* Note $Q'_{k-1}(x) = Q_{k-1}(x)$ by (7), then

$$Q'_{k-1}(y \mid s) \cdot \frac{R_i(y)}{Q'_{k-1}(y)} = Q'_{k-1}(y \mid s) \cdot \frac{Q'_{k-1}(s)}{Q'_{k-1}(s)} \cdot \frac{R_i(y)}{Q'_{k-1}(y)}$$
$$= \frac{Q'_{k-1}(y,s)}{Q_{k-1}(s)} \cdot \frac{R_i(y)}{Q'_{k-1}(y)}$$
$$= \frac{Q'_k(y,s)}{Q_{k-1}(s)}$$
$$= Q'_k(y \mid s) \cdot \frac{Q'_k(s)}{Q_{k-1}(s)}. \quad (10)$$

When $k \to \infty$, $Q'_k(x) - Q'_{k-1}(x) \to 0$. Then we have $Q'_k(s)/Q_{k-1}(s) \to 1$, and $Q'_k(y \mid s)/Q'_{k-1}(y \mid s)$. Substituting these limits into (10), we have $R_i(y)/Q'_{k-1}(y) \to 1$.

*Part 3.* Similarly, when $k \to \infty$, $Q_k(x) - Q'_k(x) \to 0$. Since $\prod_{x_j \notin y} Q'_{k-1}(x_j \mid \pi_j)$ has never changed, it can be factored out. Then according to (8) and (9), we have

$$\prod_{X_j \in Y} Q'_k(y \mid s) - Q'_k(y \mid s) \to 0,$$

$Q'_k(y \mid s)$ becomes consistent with the network structure (i.e., child – parent relations) with variables in $Y$. Since other CPT's have not been changed, $Q^*(x)$ is consistent with the structure of $N$.

*Part 4.* We prove it by an counter example. In the example in Figure 3 at the end of Section 3, an I-projection on $R_3(A,D)$ is not consistent with the network structure. The distribution $Q^*(x)$ from D-IPFP must also be consistent with the network. Since I-projection is unique, $Q^*(x)$ cannot be an I-projection of $Q_{k-1}(x)$ on $R_i(y)$. □

### 5.3 ALGORITHM D-IPFP

Now we present the algorithm of D-IPFP.

**D-IPFP**( $N_0(X)$, $R = \{R_1, R_2, \cdots R_m\}$ ) {

1. $Q_0(x) = \prod_{i=1}^n Q_0(x_i \mid \pi_i)$ where $Q_0(x_i \mid \pi_i) \in C_0$;
2. Starting with k = 1, repeat the following procedure until convergence {
   2.1. i = ((k-1) mod m) + 1;
   2.2. if $R_i(y = (x_j, z^j \subseteq \pi_j))$ /* a local constraint */
   
   { $Q_k(x_j \mid \pi_j) = Q_{k-1}(x_j \mid \pi_j) \cdot \frac{R_i(y)}{Q_{(k-1)}(y)} \cdot \alpha_k$;
   
   $Q_k(x_l \mid \pi_l) = Q_{k-1}(x_l \mid \pi_l) \quad \forall l \neq j;$ }
   
   2.3. else /* $R_i(y)$ is a non-local constraint */
   while not converge
   { $Q'_{k-1}(y \mid s) = \prod_{X_j \in Y} Q_{k-1}(x_j \mid \pi_j)$
   
   $Q'_k(y \mid s) = Q'_{k-1}(y \mid s) \cdot \frac{R_i(y)}{Q_{k-1}(y)} \cdot \alpha_k$;
   
   $\begin{cases} Q_k(x_j \mid \pi_j) = Q'_k(x_j \mid \pi_j) \forall x_j \in y \\ Q_k(x_l \mid \pi_l) = Q_{k-1}(x_l \mid \pi_l) \quad \forall x_l \notin y \end{cases};$
   }
   2.3. k = k+1;
}
3. return $N^*(X)$ with $G^* = G_0$ and $C^* = \{Q_k(x_i \mid \pi_i)\}$;
}

## 6 EXPERIMENTS

To empirically validate the algorithms and to get a sense of how expensive this approach may be, we have conducted experiments of limited scope with a artificially made network of 15 binary variables. The network structure is given in Figure 4 below.

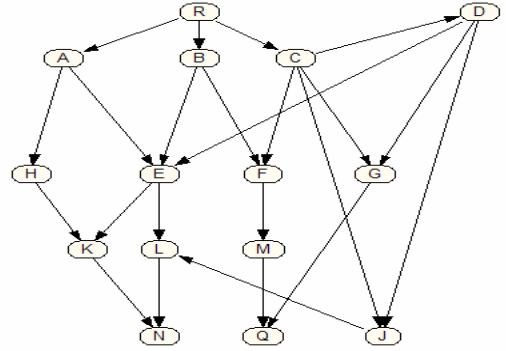

Figure 4: The Network for the Experiments

Three sets of 4, 8, and 16 constraints, respectively, are selected for the experiments. Each set contains a mix of local and non-local constraints. Number of variables in a constraint ranges from 1 to 3, the size of the subnet associated with a constraint ($|y|+|s|$) ranges from 2 to 8. Therefore a saving in computational time would be in the order of $O(2^{15-8}) = O(2^7)$. Both E-IPFP and D-IPFP were run for each of the three sets. The program is a brute force implementation of the two algorithms without any optimization. The experiments were run on a Celeron notebook CPU of 2.3G Hz and 784M maximum memory for the JVM (Java Virtual Machine). The results are given in Table 1 below.

Table 1: Experiment Results

| # of Cons. | # Iterations (E-IPFP\|D-IPFP) | | Exec. Time (E-IPFP\|D-IPFP) | | I-divergence (E-IPFP\|D-IPFP) | |
|---|---|---|---|---|---|---|
| 4 | 8 | 27 | 1264s | 1.93s | 0.08113 | 0.27492 |
| 8 | 13 | 54 | 1752s | 11.53s | 0.56442 | 0.72217 |
| 16 | 120 | 32 | 13821s | 10.20s | 2.53847 | 3.33719 |

In each iteration the program goes through all the constraints in $R$ once. Each of the 6 experimental runs converged to a distribution that satisfies all given constraints and is consistent with the network structure. As expected, D-IPFP is significantly faster than E-IPFP but with moderately larger I-divergences. The rate of speed up of D-IPFP is roughly in the theoretically estimated range ($O(2^7)$). The variation in speed up among the three sets of constraints is primarily due to the number of iterations each run takes.

## 7 CONCLUSIONS

In this paper, we developed algorithm E-IPFP that extends IPFP to modify probability distributions represented as Bayesian networks. The modification is done by only changing the conditional probability tables of the network while leaving the network structure intact. We also show a significant saving in computational cost can be achieved by decomposing the global E-IPFP into local ones with much smaller scale, as described in algorithm D-IPFP. Computer experiments of limited scope seem to validate the analysis results. These algorithms can be valuable tools in Bayesian network construction, merging and refinement when low-dimensional distributions need to be incorporated into the network.

Several pieces of existing work are particularly relevant to this work, besides those related to the development of the original IPFP and proofs of its convergence. Diaconis and Zabell (1982), in studying the role of Jeffrey's rule in updating subjective probability, consider IPFP as one of methods for mechanical updating of probability distribution. In contrast to other methods that are based on different assumptions on the subjectivity of the probabilities, the mechanical updating methods are based on some distance metrics, rather than "attempt to quantify one's new degree of belief via introspection".

Vomlel (1999) studied in detail how IPFP can be used for probabilistic knowledge integration in which a joint probability distribution (the knowledge base) is built from a set of low dimensional distributions, each of which models a sub-domain of the problem. Besides providing a cleaner, more readable convergence proof for IPFP, he also studied the behavior of IPFP with input set generated by decomposable generating class. If such input distributions can be properly ordered, IPFP may converge in one or two cycles. This kind of input set roughly corresponds to ordering constraints for a Bayesian network in such a way that the constraint involving ancestors are applied before those involving descendants, if such order can be determined. For example, if all three constraints $\{R_1(B), R_2(C), R_3(A,D)\}$ must be met, we may be better off if we apply $R_3(A,D)$ before the other two.

In all of these works, IPFP is applied to update joint distributions, none has discussed its application in modifying distribution represented by a BN.

To the best of our knowledge, the only work that applies IPFP to BN is the one by Valtorta *et al* (2000). In this work, IPFP is used to support belief update in BN by a set of soft evidences that are observed simultaneously. However, this work does not concern itself with updating the BN itself.

Algorithms developed in this paper only work with consistent constraints. It has been reported by others (Vomlel 1999, 2004) and observed by us that when constraints are inconsistent, IPFP will not converge but oscillate. How to handle inconsistent constraint is one of the important directions for future research. Another direction is to investigate in what situations modification of only conditional probability tables are no longer sufficient or desirable, the network structure need also be changed in order to better satisfy given constraints.

Efficiency of this approach also needs serious investigation. As our experiments show, IPFP in general is very expensive. The convergence time in our experiments with a small BN (15 nodes) and moderate number of constraints is in the order of hours. The performance of even D-IPFP can be bad if some input distributions involve larger number of variables. Complexity can be reduced if we can divide a large constraint into smaller ones by exploring independence between the variables (possibly based on the network structure). Properly ordering the constraints may also help. Ultimately, this problem can only be solved by parallelizing the algorithms or by approximation when the network is really large.

**Acknowledgement**

This work was supported in part by DARPA contract F30602-97-1-0215 and NSF award IIS-0326460.